# TEMPORAL REASONING WITH PROBABILITIES

Carlo BERZUINI, Riccardo BELLAZZI, Silvana QUAGLINI

*Dipartimento di Informatica e Sistemistica, Universita' di Pavia. Via Abbiategrasso 209. 27100 PAVIA (ITALY)*

## Abstract

*This paper explores the use of probabilistic graphical modelling to represent and reason about temporal knowledge. The idea is that of representing concepts and variables involved diagramatically, by means of a directed graph, called an influence diagram (ID), designed to capture probabilistic dependencies between those variables. Statistical models of progression in time, such as semi-Markov processes, can be translated into 'pieces' of influence diagram, and then embedded into large influence diagrams representing bodies of knowledge. In this way, we can include statistical modelling of time into expert systems. Stochastic simulation (Monte Carlo) approaches are proposed for probability propagation on the obtained diagrams. In particular, a combination of two techniques, known as 'Gibbs salmpling' and 'forward sampling', is discussed.*

## 1. INTRODUCTION[1]

Time plays an essential role in the modelling of reasoning processes where the involved informations are evolutive. In clinical therapy monitoring, for example, one may have to use past patient's evolutive information to predict the speed at which the patient approaches a certain critical condition, and diagnosis may involve reconstructing the patient's past progression through a set of states.

This paper explores the use of probabilistic *graphical modelling* to represent and reason about temporal knowledge. The main idea is that of representing concepts and variables involved diagramatically, by means of a directed graph. Such a graph, called an *influence diagram* (ID), is designed to capture probabilistic dependencies between variables.

Statistical models of progression in time, such as semi-Markov processes, can be translated into 'pieces' of influence diagram, and then embedded into large influence diagrams representing given pieces of knowledge. In this way, we include statistical modelling of time into expert systems. By doing so, we bring into expert systems the coherence and power of statistical models, as well as the wide accumulated statistical experience on how to 'learn' from empirical data. For what concerns semi-Markov models, for example, an extensive work on parameter estimation from partially censored data has been done (see, for example, [Lagakos 78]).

Once the appropriate ID is given, reasoning about time is a matter of propagating probabilities on it.

The complexity of influence diagrams that are typically obtained when time is involved motivated us to choose *stochastic simulation* (Monte Carlo) techniques for probability propagation. Certainly in many individual applications, especially in the simple ones, other propagation techniques would be strongly suggested (exg. those proposed in [Lauritzen 88]). But when the graphs have complex structures, for example including very large cliques such as appear when many common causal factors are modelled, there is no apparent alternative to Monte Carlo sampling. That's why this class of techniques is highly valuable within expert system shells oriented towards general applications.

Two different simulation schemes, *Gibbs sampling* [Geman 84], and *forward sampling* [Henrion 86] will be considered. Advantages of combining the two into a *composite* simulation scheme will be discussed.

Gibbs sampling has been also analyzed by [Pearl 87] and [Cooper 89] in an expert system context. These authors detected an important draw-back of Gibbs sampling, namely its unability to cope with *functional dependencies*. In this paper we tackle this problem by both trying to avoid functional dependencies in the ID design, and by combining Gibbs and forward sampling in a clever way.

## 2. MARKOV CHAINS

All of the ID's considered in this paper contain a common 'kernel', formed by a sequence of random variables $(X_0, X_1, ..., X_t)$ which take values in a finite set $S$, called the *state space*. Elements of $S$ represent possible states of the world, and are denoted $0,1,2,...,n$. Each $X_k$ is therefore a discrete random variable which takes one of $n$ possible values.

The process $X$ will always satisfy the **Markov property**:

$$p(X_{k+1}=j \mid X_0, X_1, ..., X_k) = p(X_{k+1} \mid X_k)$$

---

[1] This work was facilitated by support from MPI-40% and MPI-60% grants, and from C.N.R. grant

for all $k>0$ and $0 \leq j \leq n$. By the Markov property $X_k$ nullifies the influence that $(X_0, X_1,...,X_{k-1})$ exert on $(X_{k+1},...,X_l)$. This conditional independence pattern is depicted by the simple ID shown in Fig. 1. Such a simple ID, which forms the kernel of more complex ID's shown later, is called **Embedded Markov Chain** (EMC) of the ID. A complete probabilistic specification for the EMC is given by the following:

$$p(X_0) \qquad (1)$$
$$P_{ij} = p(X_{k+1}=j \mid X_k=i), \quad \text{for } 0 \leq i,j \leq n \qquad (2)$$

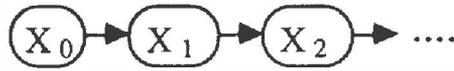

Figure 1 Embedded Markov Chain (EMC)

## 3. CONTINUOUS-TIME SEMI-MARKOV PROCESSES

Let's view the 'history' over time of a given system as a series of 'jumps' from one state of the state space to another one, ending when an 'absorbing' state is entered.

Let $X_k$ denote the state entered after the $k$-th 'jump', and let $X_0$ denote the initial state. If $X_k$ is absorbing, $X_{k+1}, X_{k+2}, X_{k+3},...$, take the conventional value '*'. In a *semi-Markov* process $X_0, X_1,...$ form a Markov chain.

Let $T_k$ denote the 'sojourn time' between the $k$-th and the $(k+1)$-th jumps, i.e. the time spent in state $X_k$. In a semi-Markov process, conditional on a given sequence of jumps, $T$'s are independent, and the distribution for $T_k$ is selected depending only on $X_k$ and $X_{k+1}$ [Ross 83].

The $X$ process and the $T$ process in a semi-Markov model can be represented by the ID shown in Fig. 2. Notice that this ID is triangulated, and all couples of parents of all nodes are joined. Therefore this ID is 'perfect', and can be interpreted in terms of conditional independence by the simple following graphical rule: 'a missing link between two nodes $A$ and $B$ indicates that $A$ is conditionally independent of $B$, given the value of nodes lying along the undirected path that connects $A$ and $B$'.

$p(X_0)$ tells us for any given state the probability that the process started from that state. Probabilities $P_{ij}=p(X_{k+1}=j \mid X_k=i)$ govern the selection of 'jumps': given that the process is in state $i$ at a given time-point, then $P_{ij}$ represents the probability that the next jump will lead it to state $j$. An additional specification, concerning the distribution of 'sojourn times' is required:

$$f_{ij} = p(T_k=t \mid X_k=i, X_{k+1}=j) \qquad (3)$$

Specifications (1)-(3) uniquely determine the joint distribution $P(X_0,X_1,X_2,... X_n,T_0,T_1,T_2,...,T_n)$ of the variables in the ID shown in Fig. 2. We refer to this property by saying that (1)-(3) specify a *Markov Random Field* (MRF) on the ID.

## 4. A SIMPLE CLINICAL EXAMPLE

The following example illustrates a general methodology, by which a temporal inference problem is solved by first designing a suitable ID and then by propagating probabilities on it. The example involves reconstructing the past patient's progression based on precise knowledge of his/her final state $s \epsilon S$, on partial information about his/her initial state $X_0$, and on the known time of transition from $X_0$ to $s$, denoted $T^{obs}$.

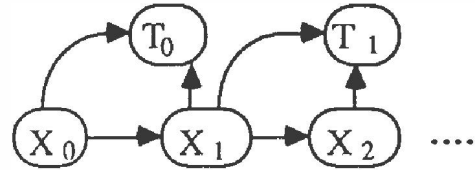

Figure 2 Simple ID for a semi-Markov process

More precisely, assume that a patient has been submitted to transplant at $t_0 = 0$. Transplant might have caused one of two viruses, $A$ and $B$, to be accidentally inoculated into the patient. In such an unlucky case, the virus would incubate, then replicate. Replication would eventually show up into a clinically detectable fever. Fever may also arise from unspecific causes unrelated to the considered viruses.

Now suppose that a given patient shows fever a period $T^{obs}=3$ months after transplant. The question is 'what's the probability that the patient has been exposed to virus $A$ ?'. Here we are only concerned in designing a suitable ID. Later we show how to use the ID for answering the question above.

There are 5 possible states: *1*- the patient has no virus and no fever, *2*- virus $A$ is incubating and no fever is present, *3*- virus $B$ is incubating and no fever is present, *4*- either virus $A$ or virus $B$ is replicating but the patient has no fever, *5*- patient's fever has been detected.

The ID for this problem is shown in Fig. 3. Nodes $X_0, X_1, X_2$ of the EMC indicate respectively the patient's initial state, the first visited state and possibly the third visited state, by taking values between 1 and 5 or the value '*'. We assume knowledge of the prior $p(X_0)$, i.e. the a priori distribution over the initial states, and of conditionals $p(X_{i+1} \mid X_i)$, $i=0,1$, that tell us to which states the patient can go once he/she is in a given state $X_i$. For example, $p(X_{i+1}=5 \mid X_i=2)$ is independent of $i$ and is the probability that a patient exposed to virus $A$ gets fever before viral replication starts, while $p(X_{i+1}=4 | X_i=1)$ must be zero since viral replication cannot occure if no virus has been inoculated.

The EMC has length 3, since it takes at most 2 'jumps' to reach the final state '5- fever' whatever the initial state. In fact, when starting from state 2 or 3 the patient first jumps to 4 and then to 5. When starting from state 1 the patient directly jumps to 5.

Let $X_0=x_0$, $X_1=x_1$ and $X_2=x_2$ denote the states visited by the patient. Let $T_0$ and $T_1$ denote times spent in $x_0$ and $x_1$ respectively. We assume $p(T_0=t | X_0,X_1) = \lambda_0 exp(-\lambda_0(t-a_0))$ with $t > a_0$, where $\lambda_0$ and $a_0$ both depend on $x_0$ and $x_1$. We assume $p(T_1=t | X_1) \lambda_1 exp(-l_1(t))$, with $\lambda_1$ depending on $x_1$. $T^{obs}$ is the observed time taken by the patient to go from state $X_0$ to state 5, so that $T^{obs}=T_0+T_1{}^u$, where $u=1$ if $X_1<5$, and zero otherwise. When $u=0$, $p(T^{obs}=t | X_0,X_1) = \lambda_0 exp(-\lambda_0(t-a_0))$ with $t > a_0$. When $u=1$, $p(T^{obs}=t | X_0,X_1) = \{\lambda_0\lambda_1 /(\lambda_0+\lambda_1)\} \{ exp(-\lambda_0(t-a_0)) - exp(-\lambda_1(t-a_0)) \}$ with $t > a_0$.

Specifications $p(X_0)$, $p(X_{i+1} | X_i)$, $i=0,1$, and $p(T^{obs}=t | X_0,X_1)$ completely specify a Markov Random Field over the ID shown in Fig. 3.

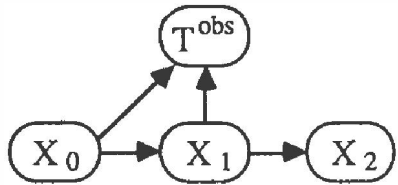

**Figure 3** ID for 'backwards' time reasoning in the 'infection' example

## 5 SIMPLIFYING THE EMC

Here we focus on the embedded Markov chain (EMC) of an ID, viewing it as a chain of binary *logical* constraints. We discuss some simple constraints propagation algorithms for simplifying the qualitative structure of the EMC *without taking into account numerical probabilities*. The simplification consists in restricting the domain of some of the $X_i$ variables in the EMC by deleting values. This has the beneficial effect of reducing the search space of algorithms that perform probabilistic reasoning.

Variables $(X_i)$, $i=1,...,n$, are discrete, since their domain is the set of states of finite-state-space Markov model. Let $v_i{}^k$ denote the $k$-th value in the domain of $X_i$, with $k=1,...,NV_i$. Given a couple of adjacent variables $X_{i-1}$ and $X_i$, a value $v_{i-1}{}^h$ of $X_{i-1}$ is **compatible** with a value $v_i{}^h$ of $X_i$ iff $P_{i-1,i} = p(X_i =v_i{}^k | X_{i-1} =v_{i-1}{}^h) >0$. The link $X_{i-1} \rightarrow X_i$ is **arc-consistent** iff for any value $v_{i-1}{}^h$ of $X_{i-1}$ there exists at least a value $v_i{}^k$ of $X_i$ which is compatible with $v_{i-1}{}^k$. The notion of arc-consistency has been introduced by [Montanari 74] and [Mackworth 77]. The following algorithm deletes values from the domain of $X_{i-1}$ until the directed arc $X_{i-1} \rightarrow X_i$ is arc-consistent :

REVISE_L (i)
1) For $h=1,...,NV_{i-1}$ do
2)     if $P_{i-1,i} = p(X_i =v_i{}^k | X_{i-1} = v_{i-1}{}^h) = 0$ for $k=1,...,NV_i$ then
3)     delete $v_{i-1}{}^h$ from the domain of $X_{i-1}$.

The EMC of an ID is arc-consistent if all $X_{i-1} \rightarrow X_i$ arcs are arc-consistent. The following algorithm makes the EMC arc-consistent by revising each of its links:

REVISE_G
1) Until there is no change do
2)     for $i=2$ to $n$ by 1 do
3)         perform REVISE_L (i)
4) end.

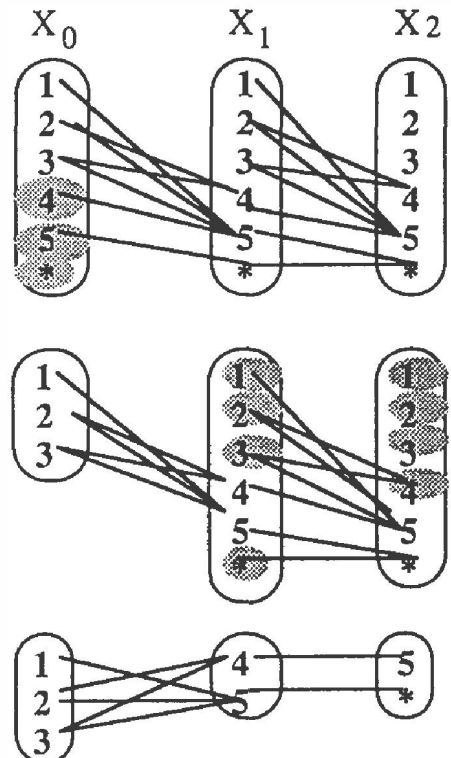

**Figure 4** Deletion of excluded initial states and revision of the EMC of the ID for the 'infection' example.

A MG simplified by means of algorithm REVISE_G is said to be revised.

Given an inference problem on a Markov process, we may often *a priori* exclude the possibility that certain variables $X_i$ of the EMC, most often the initial state $X_0$, take certain values. We may then delete such values from the domain of those $X_i$, and then revise the ID. Revision propagates the deletion in such a way that the domain of other $X_i$ may be reduced, too. The global revision procedure is the following algorithm.

```
GLOBAL
1)   begin
2)       for i=1 to n by 1 do
3)           delete from the domain of X_i those
values            which are a priori excluded
4)           perform REVISE_G
5)       end
6)   end.
```

Fig.4 shows a sequence of deletion and revision steps performed on the EMC of the ID introduced for the 'infection' example. In this figure the domains of variables in the EMC are visible, and compatibility relations across these domains are indicated by undirected links. The first step is deleting $X_0=4$, $X_0=5$ and $X_0=*$ since they can be excluded as initial states. Then revision is performed.

# 6. REASONING

## 6.1 A sampling approach

Since an ID with associated MRF constitutes a complete probabilistic model of the variables included, it contains the information needed to answer all probabilistic queries about these variables. These queries might be requests of :

(i) *prediction* = computing the degree of belief of future states of the system given its present and past states,
(ii) *explanation* = computing the degree of belief of past states of the system from its present state,
(iii) *reconstruction* = computing the degree of belief of system states during $(t_0,t_1)$ given incomplete information on the state of the system at $t_0$ and at $t_1$.

A special case of prediction is predicting the consequences of a set of alternative action plans; if utility information is provided, a recommendation of the 'best' action plan may be obtained.

Let the variables in the ID be uniformly denoted $Y_i$ $i=1,...,n$. Evidence, such as observing certain states or transitions at certain times, is translated into an assignment of values (instantiation) to a subset of $\{Y_1, Y_2, ... ,Y_n\}$. Without loss of generality, we take this subset to include the last $(n-p)$ variables of the list, $\{Y_{p+1},Y_{p+2} ..... ,Y_n\}$, and denote by $y_{p+1}, y_{p+2},...,y_n$ respective assigned values. $Y_1,Y_2,...Y_p$ are called *free* variables. Since any potential 'history' of the system corresponds to a configuration of values of free variables $\{Y_1=y_1,Y_2=y_2,...Y_p=y_p\}$, the degree of belief into a given history is formally the posterior density $P(Y_1=y_1,Y_2=y_2,...Y_p=y_p | y_{p+1},y_{p+2}, ..., y_n )$.

The method described in following sections indirectly samples $P(Y_1,Y_2,...,Y_p | y_{p+1},y_{p+2},...,y_n)$ by exploiting the conditionals $P(Y_i | P_i)$. Once an appropriate sample has been drawn, we can use it to estimate the desired posteriors $P(Y_1=y_1,Y_2=y_2,...Y_p=y_p | y_{p+1},y_{p+2},...,y_n )$.

## 6.2 General sampling strategy

In order to compute posterior probabilities, we may adopt the method of counting how frequently free variables take certain patterns of values in a series of simulation runs. We can use the ID and probability specifications attached to it to generate random samples of hypothetical histories that are likely to occur (or have occurred) at the light of available information.

We discuss below two different mechanisms to generate random samples from an ID, namely *Gibbs sampling* and *forward sampling*. They have a basic scheme in common: each node $Y$ in the ID is viewed as a separate processor, which receives control once in each simulation cycle. Each time it receives control, the generic $Y$ computes the conditional distribution for the hosted variable given the states of *neighbouring variables*, then samples the computed distribution and instantiates the hosted variable to the value selected by the sampling. The cycle repeats itself by sequentially 'visiting' all the nodes in the ID. Each cycle over the set of ID nodes yields one simulated 'history'. From a set of simulated histories one may compute required marginal a posteriori probabilities.

The length of computation of sampling algorithms is determined mainly by the degree of accuracy desired, not by the topology of the dependencies embodied in the ID, and increases only linearly with the number of nodes. They have an inherent *parallelism*: processors associated with variables may be activated concurrently provided that no two adjacent processors are activated at the same time.

Gibbs sampling and forward sampling differ as far as the visiting order and the definition of 'neighbourhood' are concerned.

## 6.3 Gibbs sampling

In Gibbs sampling, the nodes are visited in an arbitrary order, and the "neighbourhood" $w_Y$ of the generic node $Y$ is defined to include:
(i) its parents, $P_Y$ ,

17

*(ii)* its children, $C_Y$, and

*(iii)* $R_Y$, the set of all parents of $C_Y$, except $Y$.

Then the distribution for $Y$ given $w_Y$ is computed as [Pearl 87]:

$$p(y|w_Y) = \text{const } p(y|P_Y) \, p(C_Y|y, R_Y) \qquad (4)$$

Given an arbitrary starting set of values $(y^{(0)}_1, y^{(0)}_2, \ldots, y^{(0)}_p)$, the algorithm cycles over the set of free variables. After $h$ such iterations we would arrive at $(y^{(h)}_1, y^{(h)}_2, \ldots, y^{(h)}_p)$. [Geman 84] shows that under mild conditions convergence is ensured in the sense that $(y^{(h)}_1, y^{(h)}_2, \ldots, y^{(h)}_p) \sim P(Y_1, Y_2, \ldots, Y_p | y_{p+1}, y_{p+2}, \ldots, y_n)$ as $h \to \infty$. Thus, for $h$ large enough, we can regard $(y^{(h)}_1, y^{(h)}_2, \ldots, y^{(h)}_p)$ as a 'history' drawn from $P(Y_1, Y_2, \ldots Y_p | y_{p+1}, y_{p+2}, \ldots, y_n)$. In conclusion, given an ID with $p$ free variables, and assuming that convergence is guaranteed, the above Gibbs sampling scheme requires $ph$ random variate generations to yield one simulated 'history'.

The generic iteration of the Gibbs sampling produces a new configuration $(y^{(h+1)}_1, y^{(h+1)}_2, \ldots, y^{(h+1)}_p)$ from the current $(y^{(h)}_1, y^{(h)}_2, \ldots, y^{(h)}_p)$. For convergence to be guaranteed, it is required that for any pair $(i,j)$ of conceivable configurations of the ID there is a positive probability of reaching $j$ from $i$ in a finite number of Gibbs sampling iterations. This is called *reachability condition*.

In the very special case in which the ID is as simple as a Markov chain it's easy to check the reachability condition. Considering for example Figs. 6(a) and (b), it is evident that the elementary Markov chain in (a) satisfies the reachability condition, while the one in (b) does not. To see this, consider for example that starting Gibbs sampling on the ID in Fig. 6(a) from configuration (3,3) there is a non-null probability for the next iteration to generate (2,2) and for the subsequent one to generate (1,1). This is not true in Fig. 6(b), since when $X_1$ is equal to 3 $X_2$ is locked to value 2 or 3, and conversely when $X_2$ has value 2 or 3 $X_1$ is locked to value 3. It is easy to see that the crucial feature that distinguishes the two examples is the connectedness of the graph of compatibility between the values of the two nodes. Such a graph is completely connected in Fig. 5(a), but not in Fig. 5(b).

It is straightforward to generalise this by stating that convergence of the Gibbs sampling algorithm is guaranteed on a discrete Markov chain *iff* all pairs of adjacent nodes are completely connected. That is: *local* connectivity in the chain ensures *global* connectivity in the space of configurations of values of the chain. In real-world applications this is often not the case (see for example the 'infection' example).

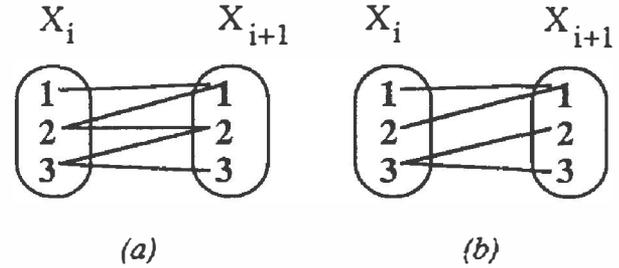

**Figure 5** *(a)* completely connected nodes, *(b)* non-completely connected nodes.

Checking the reachability condition on ID's with general structure may be markedly more difficult than we have seen on a Markov chain, especially in presence of continuous variables. Convergence of Gibbs sampling may fall down to zero when: *(i)* there are variables constrained by functional dependencies, as for example $X = Y^2$, or *(ii)* a continuous variable selects its distribution depending on the value of parent nodes, and selectable distributions have dijoint supports.

A practical solution to the problems above consists in: *(a)* trying to avoid functional dependencies in the design of the model, and *(b)* combining the Gibbs sampling scheme with a *different scheme, called forward sampling*, whose convergence is not affected by connectivity features. This scheme is described in the next subsection.

### 6.4 Forward sampling

In 'forward' sampling, a scheme proposed by [Henrion 86], free variables are visited according to the order of 'causation' in the ID. Orphan nodes and nodes whose parents are all instantiated are visited first. Orphan nodes sample prior distributions associated to their respective hosted variables. The generic non-orphan node $Y$ samples $P(Y|P_Y)$, ignoring the values of neighbours other than its parents. This propagation scheme from causes to effects may produce inconsistent samples. In fact, in those cases when some observable nodes are non-orphan, nothing prevents forward sampling from the possibility of assigning a free variable $Y$, parent of an instantiated variable $X$, a value $y$ which clashes with the neighbouring value $x$. It may take many runs before a consistent sample is generated.

### 6.5 Composite sampling

Since Gibbs and forward sampling suffer from independent drawbacks, their complementary use may be advantageous. Here is a sensible combination of the two techniques:

(1) run 'forward' sampling until $m$ consistent samples are obtained;
(2) use each of the above $m$ samples as starting set for $h$ iterations of the Gibbs sampler;
(3) $m$ simulated 'histories' $(Y^{(h)}_{1i}, Y^{(h)}_{2i},...,Y^{(h)}_{pi})$, $i=1,...,m$ are obtained through step (2);
(4) the degree of belief associated to a given hypothetical history may be computed by the *'kernel' density estimator*, i.e. as the fraction of the $m$ final sampled histories that match the hypothesized one.

Parameters $m$ and $h$ should be tuned depending on the application at hand. As a rule of thumb, in those cases where all instantiated nodes are orphan, one could take $h=0$. When struggling against slow convergence rate of Gibbs sampling one should try increasing $m$ and decreasing $h$.

### 6.6 Illustration

With reference to the 'infection' example, the answer to the query *'was the patient exposed to virus A ?'* is the posterior $p(X_0=2 \mid T^{obs}=3 months)$, which we compute below using the composite sampling scheme.

First, 'forward-generate' $m$ consistent histories $(x_{0i}, x_{1i}, x_{1i})$, $i=1,...,m$. This is done by repeating the following sequence until $m$ histories are obtained ($\sim$ means 'drawn from') :

(1) $x_0 \sim p(X_0 = x_0)$,
(2) $x_1 \sim p(X_1 = x_1 \mid X_0 = x_0)$
(3) $x_2 \sim p(X_2 = x_2 \mid X_1 = x_1)$
(4) reject if $p(T^{obs}=3months \mid X_0=x_0, X_1=x_1) = 0$

Second, use each of the 'forward-generated' histories $(x_{0i}, x_{1i}, x_{1i})$, $i=1,...,m$, as a starting set for $h$ Gibbs sampling cycles:

$x^{(1)}_0 \sim p(X_0 = x^{(1)}_0 \mid X_1=x_1, T^{obs}=3) = p(X_0 = x^{(1)}_0)$
$\quad p(X_1=x_1 \mid X_0=x^{(1)}_0) p(T^{obs}=3 \mid X_0=x^{(1)}_0, X_1=x_1)$

$x^{(1)}_1 \sim p(X_1 = x^{(1)}_1 \mid X_0 = x^{(1)}_0, T^{obs}=3, X_2=x_2)$
$= p(X_1=x^{(1)}_1 \mid X_0=x^{(1)}_0) p(T^{obs}=3 \mid X_0=x^{(1)}_0, X_1=x^{(1)}_1) p(X_2=x_2 \mid X_1=x^{(1)}_1)$

$x^{(1)}_2 \sim p(X_2 = x^{(1)}_2 \mid X_1=x^{(1)}_1)$

$x^{(2)}_0 \sim p(X_0 = x^{(2)}_0 \mid X_1 = x^{(1)}_1, T_0 = t^{(1)}_0) =$
$p(X_0=x^{(2)}_0) p(X_1=x^{(1)}_1 \mid X_0=x^{(2)}_0) p(T^{obs}=3 \mid X_0=x^{(2)}_0, X_1=x^{(1)}_1)$

$x^{(2)}_1 \sim p(X_1=x^{(2)}_1 \mid X_0=x^{(2)}_0, T^{obs}=3, X_2=x^{(1)}_2)$
$x^{(2)}_2 \sim p(X_2 = x^{(2)}_2 \mid X_1=x^{(2)}_1)$
. .
. .
$x^{(h)}_2 \sim p(X_2 = x^{(h)}_2 \mid X_1=x^{(h)}_1)$

At each of the $m$ cycles we save the final $(x^{(h)}_0, x^{(h)}_1, x^{(h)}_2)$. We end with a set of $m$ 3-uples $(x^{(h)}_{0i}, x^{(h)}_{1i}, x^{(h)}_{2i})$, $i=1,...,m$.

Third, apply the *finite mixture density estimator* [Gelfand 88]:

$p(X_0 = 2 \mid T^{obs}=3) = m^{-1} \sum_{i=1}^{m} P(X_0 = 2 \mid X_1 = x^{(h)}_{1i}, T^{obs}=3)$

whose efficiency is superior to the usual *kernel density estimator*, which would count the fraction of times $X_0$ takes value 2 among the $m$ simulated histories.

## 7. MODELLING THE LIFE-THREAT DUE TO DRUG-INDUCED TOXICITY IN THE TREATMENT OF CANCER.

The methodology outlined in the previous sections allows constructing statistical models of non-trivial decision processes. Within an expert system, models of such a kind may provide the machinery for solving important decisions.

The skillful modeller must have two fundamental varieties of competence. First, he/she must have a substantial knowledge of the decision domain, including a perception of which are the crucial problems to be solved, and a causal understanding about factors that govern or influence temporal processes of change. Second, he/she must be able to recognize and name all statistical entities involved in the model, and must know how to relate and interlock them to form the complete influence diagram.

These varieties of competence are shown in action in the example discussed below, which concerns a clinical application in the field of cancer treatment. Due to lack of space, we shall avoid most of the technical details.

One of the main issues in the treatment of cancer concerns balancing the life threat posed by administering a given treatment against the benefits that the treatment is likely to give. In general, the drugs used for cancer chemotherapy give rise to toxicity, which is often associated with sites where cells are self-renewing. In the following we restrict to modelling toxicity in a single site, say the bone marrow.

The behaviour of the expert clinician in front of the decision problems posed by such a trade-off is often described by means of 'condition->action' rules which do not explicitly reveal the action motivations in terms of the effects on survival probabilities. Designing an ID model of the life-threatening aspects of toxicity may pose a remedy to such a deficiency, help clarifying how clinicians view toxicity and how their treatment is influenced by their perception of risk of toxicity. It may also be used as an operational tool for treatment advice. The quantities involved in the model and the physiological assumptions embodied in it have been chosen following [Gallivan 88].

The 'kernel' of the model is a sequence of states $R_i$ of the bone marrow progressing through regularly spaced times $i=0, 1, ..., n$, where $R_i$ is a continuous variable taking values in $(0,w)$ corresponding to increasing levels of dysfunction of the patient's bone marrow. The random evolution of bone marrow states is described by an autoregressive scheme :

$$r_i = \alpha_0 + r_{i-1}(\alpha_1 + \alpha_2 d_i) + \varepsilon_i$$

where the $\varepsilon_i$ are independent random disturbances with assumed known distributions $No(0, \sigma^2)$, for all $i > 0$, $d_i$ is the dose level administered between time $i-1$ and time $i$, and $\alpha = (\alpha_0\ \alpha_1\ \alpha_2)'$ is a vector of unknown parameters, with known prior $p(\alpha)$.

In particular, parameter $\alpha_2$ models the toxic effect exerted upon the bone marrow by the administered drug. Assuming an 'all or none' drug administration regimen, we restrict the variable $D_i$ to take value 1 if during the $i$-th interval the drug is administered, and value 0 otherwise.

Parallel to the $R_i$ sequence, there is a sequence $X_i$ of global states of the patient. $X_i = 0$ means that at time $i$ the patient is dead, $X_i = 1$ means that at time $i$ the patient is alive. $X_0, X_1,...$ is modelled as a discrete-time Markov chain with one-step transition probabilities 'modulated' by the level $r_i$ of dysfunction of the bone marrow. More precisely :

$p(X_i = 'alive' \mid X_{i-1} = 'alive', r_{i-1}) = exp(-kT(1- s(r_{i-1})))$

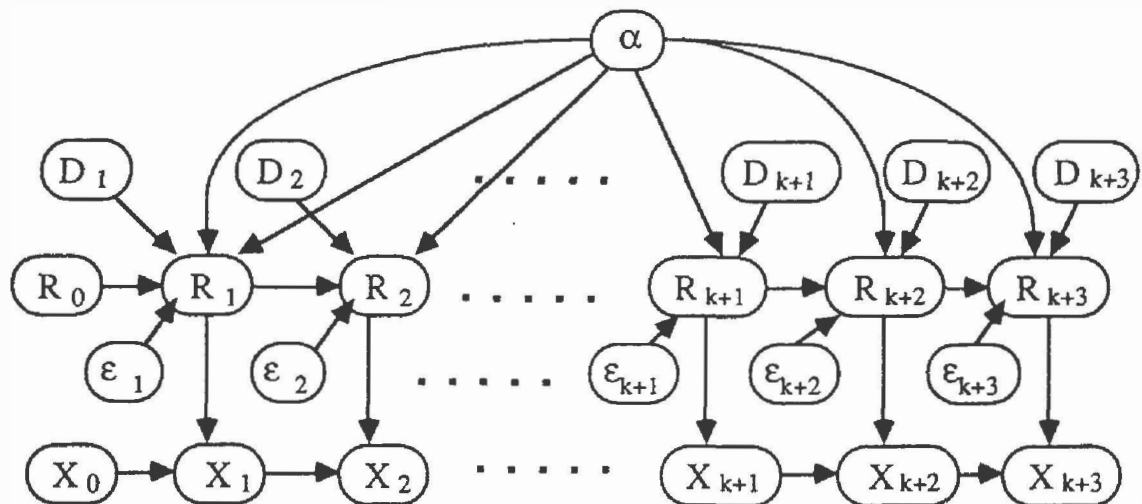

**Figure 6** ID for therapy monitoring in cancer.

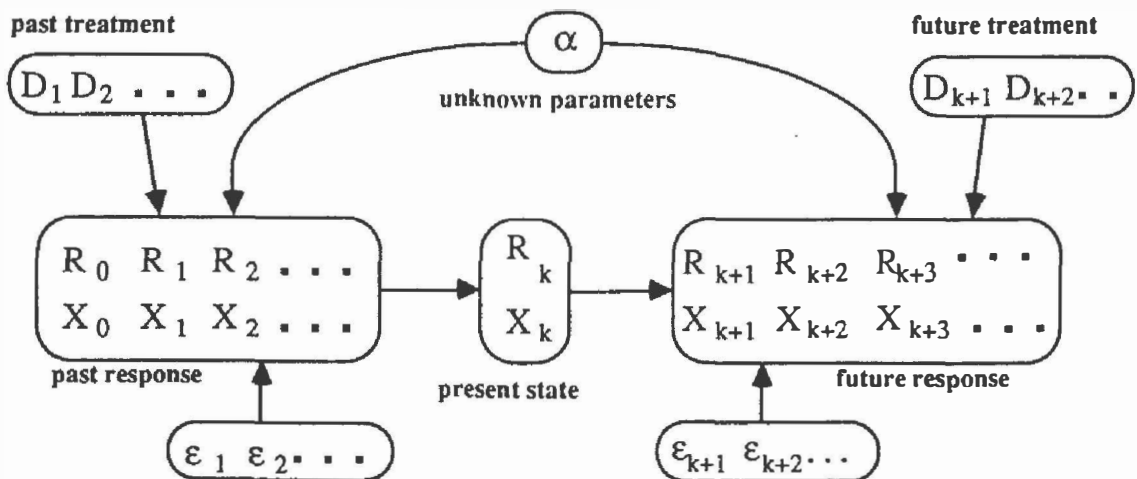

**Figure 7** Schematic re-expression of the ID shown in Fig. 6.

20

for all $i=1,...,n$, where $s(r_{i-1})$ is the known probability of surviving an infection with a bone marrow at level $r_{i-1}$ of dysfunction, $T$ is the length of the time step of the Markov chain, and $k$ is the known average rate at which bouts of infection hit the patient.

The quantities above and the conditional independence relationships implicit in the specifications above are translated into the ID shown in Fig.6. The variables correspond to the nodes of the ID and the missing links of the ID have the meaning of conditional independencies. The probabilistic knowledge described by the distributional assumptions specified above is stored in the ID nodes. Each node $X_i$, $i=1,...,n$, for example, stores probabilities $p(X_i = \text{'alive'} | X_{i-1} = \text{'alive'}, r_{i-1})$ and $p(X_i = \text{'alive'} | X_{i-1} = \text{'dead'}, r_{i-1})$.

We are now ready to put the ID into operation, to perform required inferences. Both *retrospective* learning and *prospective* inference is featured on the ID. Knowledge about the past patient's response up to time $k$ is used to learn the posterior of $\alpha$, and this posterior is used to predict the chances of patient's survival under a given treatment at future times $k+1, k+2,...$.

The schematic ID shown in Fig. 7 helps clarifying the inferencing scheme. $D_i$, $i=1,...,k$, $X_i$ and $R_i$, $i=0,...,k$ are assumed observables, since they represent the past treatment and the observed response to it, respectively. The posterior of $\alpha$ is derived by propagating probabilities from these observables. Then, we condition on $D_i$, $i=k+1, k+2,...$, (future treatment), and, from the current patient's state $X_k, R_k$, we infer $X_i$, $i=k+1, k+2,...$, taking into account the posterior of $\alpha$.

## ACKNOWLEDGEMENTS

The authors are indebted to M.Leaning, J.Pearl, D.Spiegelhalter and M. Stefanelli for stimulating discussions on the topic. The authors' insight into the problem was greatly helped by participation in the 1989 Edinburgh Workshop on Statistics and Expert Systems, supported by the U.K. SERC and organized by Drs. D.Hand, D.Spiegelhalter and Mr. P.R.Fisk.